\documentclass[10pt,twocolumn,letterpaper]{article}

\usepackage[pagenumbers]{cvpr} 
\usepackage[numbers,sort]{natbib}
\usepackage{graphicx}
\usepackage{amsmath}
\usepackage{amssymb}
\usepackage{booktabs}
\usepackage{algorithm}
\usepackage{lipsum}
\usepackage{url}

\usepackage{mathtools}

\usepackage{amsfonts}
\usepackage{array}
\usepackage{microtype}
\usepackage{enumitem}
\usepackage{mathtools}
\usepackage{makecell}
\usepackage{colortbl}
\usepackage{pifont}
\usepackage{tabularx}
\usepackage{multirow}

\usepackage{trimclip}

\usepackage{listings}
\usepackage{upquote}  
\usepackage{xcolor,colortbl}
\lstdefinestyle{overleaf}{
    backgroundcolor=\color[rgb]{0.95,0.95,0.92},   
    commentstyle=\color[rgb]{0,0.6,0},
    keywordstyle=\color{magenta},
    numberstyle=\tiny\color[rgb]{0.5,0.5,0.5},
    stringstyle=\color[rgb]{0.58,0,0.82},
    basicstyle=\ttfamily\footnotesize,
    breakatwhitespace=false,         
    breaklines=true,                 
    captionpos=b,                    
    keepspaces=true,                 
    numbers=left,                    
    numbersep=5pt,                  
    showspaces=false,                
    showstringspaces=false,
    showtabs=false,                  
    tabsize=2
}
\lstdefinestyle{mocov3}{
  backgroundcolor=\color{white},
  basicstyle=\fontsize{7.5pt}{7.5pt}\ttfamily\selectfont,
  columns=fullflexible,
  breaklines=true,
  captionpos=b,
  commentstyle=\fontsize{7.5pt}{7.5pt}\color[rgb]{0.25,0.5,0.5},
  keywordstyle=\fontsize{7.5pt}{7.5pt}\color[rgb]{0.85,0.18,0.50},
}
\usepackage{etoolbox}
\makeatletter
\AfterEndEnvironment{algorithm}{\let\@algcomment\relax}
\AtEndEnvironment{algorithm}{\kern2pt\hrule\relax\vskip3pt\@algcomment}
\let\@algcomment\relax
\newcommand\algcomment[1]{\def\@algcomment{\footnotesize#1}}
\renewcommand\fs@ruled{\def\@fs@cfont{\bfseries}\let\@fs@capt\floatc@ruled
  \def\@fs@pre{\hrule height.8pt depth0pt \kern2pt}%
  \def\@fs@post{}%
  \def\@fs@mid{\kern2pt\hrule\kern2pt}%
  \let\@fs@iftopcapt\iftrue}
\makeatother

\usepackage[pagebackref,breaklinks,colorlinks]{hyperref}
\usepackage[capitalize]{cleveref}
\crefname{section}{Sec.}{Secs.}
\Crefname{section}{Section}{Sections}
\Crefname{table}{Table}{Tables}
\crefname{table}{Tab.}{Tabs.}

\def\cvprPaperID{0000} 
\def\confName{CVPR}
\def\confYear{2023}

\newcommand{\authsep}{\;\;}

\begin{document}

\title{CariesBench: A Hierarchical Multi-view Benchmark for Dental Caries Screening and Diagnostic Reasoning from 3D Intraoral Scans}

\author{Xuefen Liu$^1$ \authsep Yanxiong Li$^2$ \authsep Xinquan Yang$^1$ \authsep Fangqian Zhang$^2$ \authsep Qinyi Lu$^2$ \\
Yuning Zhang$^2$ \authsep Xianxu Hou$^3$ \authsep Shinan Zhang$^{2\star}$ \authsep  Wen Ma$^{2\star}$ \authsep Linlin Shen$^{1\star}$ \\
}

\maketitle
{\let\thefootnote\relax\footnote{
{$^{\star}$Corresponding author. \\ 
$^{\quad\;\;\ 1}$ Shenzhen University.\\
$^{\quad\;\;\ 2}$ Kunming Medical University.\\
$^{\quad\;\;\ 3}$ Xi'an Jiaotong-Liverpool University.}}}

\begin{abstract}
Dental caries assessment is a structured clinical process requiring lesion localization, identification of the affected tooth and surface, and severity grading rather than simple image-level recognition. 3D intraoral scans (IOS) preserve detailed dental geometry and surface morphology and support standardized rendering from complementary viewpoints, enabling fine-grained visual assessment while introducing a distinct need for anatomically grounded multi-view reasoning. Despite recent advances in multimodal large language models (MLLMs) and dental vision-language benchmarks, existing evaluations do not jointly assess hierarchical arch-to-tooth-to-surface reasoning, cross-view evidence integration, and surface-level caries assessment. To address this gap, we introduce CariesBench, a hierarchical benchmark constructed from 5,932 3D IOS, comprising 85,455 standardized multi-view renderings and 133,994 clinically grounded visual question answering instances with expert quality control. CariesBench comprises three complementary task families: Arch-level Clinical Screening (ACS), Tooth-level Diagnostic Reasoning (TDR), and Structured Clinical Decision-Making (SCD). Using the CariesBench training set, we further fine-tune Qwen3-VL-8B-Instruct to obtain IntraoralGPT. Extensive evaluation of 24 representative baseline MLLMs reveals a substantial benchmark-level performance disparity between arch-level screening and tooth-level diagnostic reasoning, with higher accuracy on ACS than on TDR. CariesBench-specific supervision substantially improves the Qwen3-VL-8B-Instruct baseline, increasing the 19-task macro-average clinical accuracy from 52.7\% to 69.5\% and TDR Overall from 34.4\% to 44.7\%. Despite these gains, fine-grained TDR remains challenging, with controlled ablations revealing task-dependent benefits from complementary visual observations. CariesBench thus provides a clinically grounded testbed for evaluating hierarchical multi-view reasoning in digital dentistry.

\end{abstract}


\section{Introduction}
\begin{figure*}[ht]
    \centering
    \includegraphics[width=\linewidth]{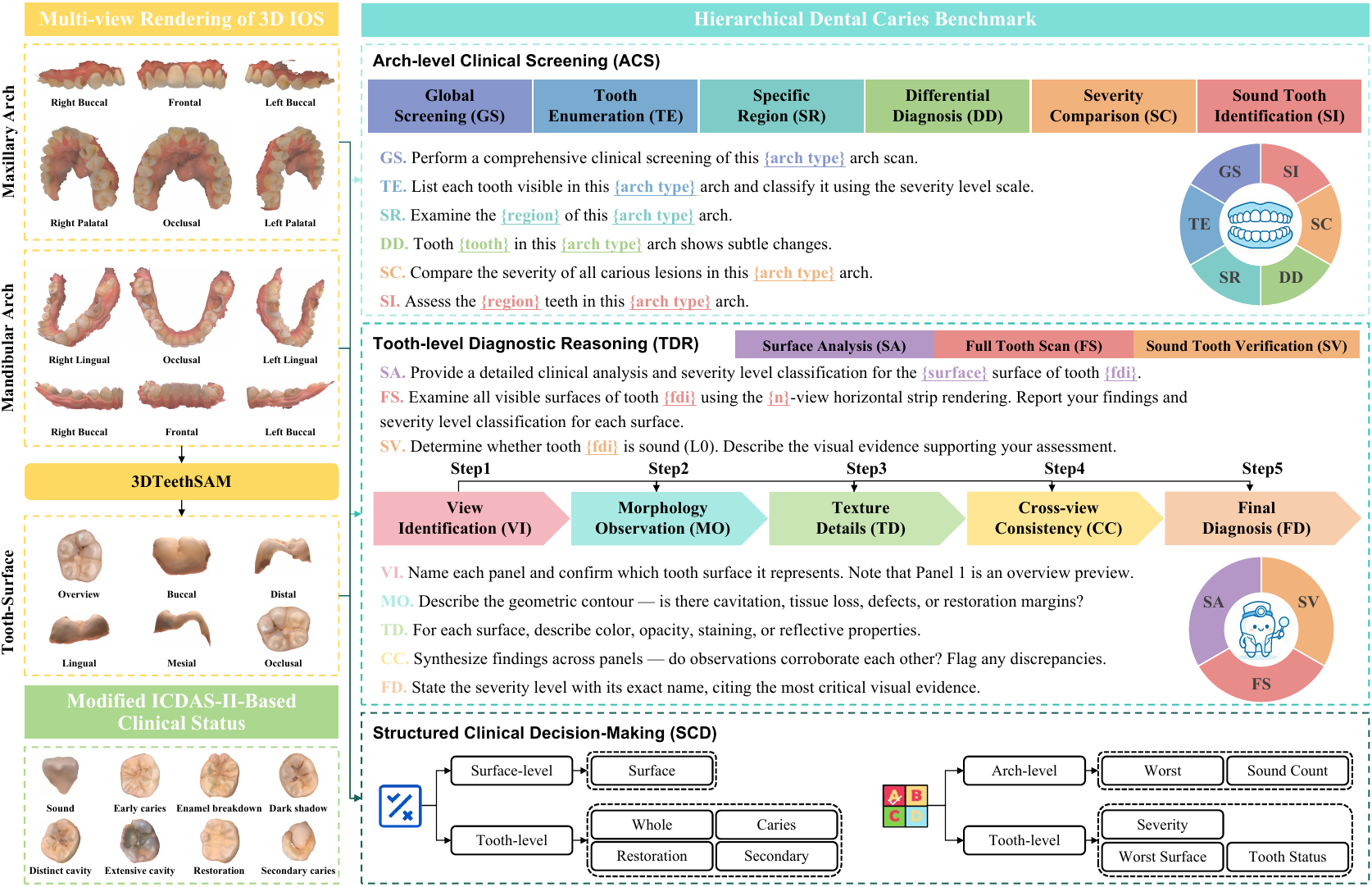}
    \caption{Overview of CariesBench. Starting from complete 3D intraoral scans, standardized arch-level rendering and tooth-level segmentation produce anatomically aligned multi-view representations. CariesBench organizes evaluation hierarchically into Arch-level Clinical Screening (ACS), Tooth-level Diagnostic Reasoning (TDR), and Structured Clinical Decision-Making (SCD), with TDR incorporating a five-step progressive reasoning chain. The benchmark comprises 5,932 3D IOS, 85,455 multi-view images, and 133,994 quality-controlled VQA instances.
}
    \label{fig:overview}
\end{figure*}

Dental caries is a common oral disease whose clinical management depends on accurate lesion localization and severity assessment. Caries diagnosis is therefore not merely a binary recognition task. Clinicians must identify the affected tooth and surface and grade lesion severity according to standardized diagnostic criteria. The International Caries Detection and Assessment System (ICDAS)~\cite{ismail2007international,dikmen2015icdas}, for example, defines caries status at the tooth-surface level and stratifies lesion severity according to visible structural changes. Reliable caries assessment therefore requires visual evidence to be anatomically localized and interpreted at the appropriate clinical severity, making the task inherently tooth-specific, surface-aware, and fine-grained.

3D intraoral scans (IOS) provide a particularly informative representation for such assessment. By preserving detailed dental geometry and surface morphology, IOS enable individual teeth and surfaces to be examined from complementary viewpoints, with previous clinical studies demonstrating their potential for detecting and classifying carious lesions at tooth and surface levels~\citep{ntovas2023occlusal,eggmann2024recent}. However, this richer representation also introduces a distinct reasoning challenge: clinically relevant evidence may be distributed across complementary views of the same anatomy. Reliable interpretation therefore requires localizing findings to the correct tooth and surface, maintaining anatomical correspondence, and integrating evidence across views before severity assessment. IOS-based caries assessment is therefore not merely a single-image recognition problem, but an \emph{anatomically grounded multi-view reasoning} problem.

Recent advances in multimodal large language models (MLLMs) have demonstrated strong capabilities in medical visual understanding and reasoning across diverse imaging modalities and clinical tasks~\citep{he2020pathvqa,liu2021slake,hu2024omnimedvqa,li2023llava}, motivating growing research interest in dentistry. Dental datasets and benchmarks have accordingly evolved from 2D pathology recognition and localization toward visual question answering, anatomical reasoning, and multimodal interpretation. Traditional resources such as DENTEX~\citep{hamamci2023dentex} primarily focus on recognition or localization from 2D dental images, whereas recent vision-language benchmarks extend to dental VQA~\citep{li2026metadent}, arch- and tooth-level understanding~\citep{hao2026towards}, multimodal dental reasoning~\citep{meng2026multimodal}, and native 3D IOS modeling~\citep{xiong2026iosvlm}. Nevertheless, these benchmarks individually cover only subsets of the combined capabilities targeted by IOS-based caries assessment. As summarized in Table~\ref{tab:benchmark_comparison}, none of the compared resources jointly evaluates hierarchical arch-to-tooth-to-surface reasoning, anatomically corresponding multi-view evidence integration, and surface-level caries assessment within a unified benchmark. Consequently, it remains unclear whether current MLLMs can maintain anatomical correspondence and integrate complementary evidence as diagnosis progresses from global screening to localized tooth- and surface-level assessment.

To address this gap, we introduce CariesBench, a hierarchical benchmark for evaluating clinically grounded multi-view reasoning over observations derived from 3D IOS. Constructed from 5,932 3D IOS, CariesBench comprises 85,455 standardized multi-view 2D renderings and 133,994 clinically grounded VQA instances. A stratified 20\% subset of benchmark instances underwent expert validation by two dentists with 8 and 3 years of clinical experience, respectively, under a standardized protocol designed and supervised by a senior dentist with 15 years of clinical experience; disagreements were resolved through adjudication under senior clinical supervision. Unlike conventional image-level VQA, CariesBench derives complementary observations from the same underlying 3D anatomy, thereby preserving correspondence across views. Evaluation spans three complementary task families: Arch-level Clinical Screening (ACS) for global dentition-level understanding, Tooth-level Diagnostic Reasoning (TDR) for localized reasoning over individual teeth and surfaces, and Structured Clinical Decision-Making (SCD) for objectively scored clinical decisions across anatomical granularities. This hierarchy enables systematic evaluation of MLLMs as the diagnostic scope narrows from the entire arch to individual teeth and surfaces.

Extensive evaluation of 24 representative baseline MLLMs reveals a substantial benchmark-level performance disparity between arch-level screening and localized tooth-level diagnostic tasks. Across model families, performance is consistently higher on arch-level screening than on localized tooth-level diagnosis, indicating an observed global-to-local performance disparity across the evaluated tasks. The lowest performance is observed on TDR tasks involving fine-grained anatomical localization and surface-level clinical-status prediction. For example, GPT-5.5 achieves 76.9\% on Sound Tooth Identification and 65.9\% on Tooth Enumeration in ACS, but only 31.8\% on Surface Analysis and 30.5\% on Full Tooth Scan in TDR. These results highlight the difficulty of localized diagnosis despite strong arch-level performance.

To investigate whether these limitations can be mitigated through benchmark-specific supervision, we fine-tune Qwen3-VL-8B-Instruct on the CariesBench training set to obtain IntraoralGPT. Under the same patient-level held-out protocol, IntraoralGPT improves the 19-task macro-average clinical accuracy of Qwen3-VL-8B-Instruct from 52.7\% to 69.5\% (+16.8 points). On TDR, its overall accuracy increases from 34.4\% to 44.7\%, with Surface Analysis improving from 24.3\% to 30.6\% and Full Tooth Scan from 23.0\% to 29.1\%. These gains demonstrate the value of CariesBench-specific supervision for hierarchical intraoral reasoning; however, the resulting 44.7\% TDR Overall remains slightly below the strongest zero-shot baseline of 45.9\%, underscoring the continued difficulty of fine-grained anatomical localization and cross-view evidence integration.

To summarize, our contributions are threefold:
\begin{enumerate}[label=\arabic*), leftmargin=*]
\item We introduce CariesBench, a large-scale benchmark built from 5,932 3D intraoral scans, 85,455 standardized multi-view renderings, and 133,994 clinically grounded VQA instances, enabling hierarchical evaluation from arch-level screening to tooth- and surface-level reasoning over anatomically corresponding views. 
\item We comprehensively benchmark 24 representative MLLMs and identify a substantial performance disparity between arch-level screening and tooth-level diagnostic tasks, while providing controlled within-task evidence on the task-dependent contribution of complementary visual observations.
\item We introduce IntraoralGPT, obtained by fine-tuning Qwen3-VL-8B-Instruct with CariesBench supervision, and demonstrate substantial improvements in tooth-level multi-view reasoning and structured clinical decision-making.
\end{enumerate}

\section{Related Work}

\begin{table*}[ht]
\centering
\setlength{\tabcolsep}{3pt}
\renewcommand{\arraystretch}{1}
\caption{
Comparison of representative medical and dental multimodal datasets and VQA benchmarks.
CariesBench uniquely combines hierarchical anatomical reasoning
(Arch $\rightarrow$ Tooth $\rightarrow$ Surface), multi-view IOS observations,
and explicit diagnostic reasoning among the compared resources.
}
\label{tab:benchmark_comparison}

\scriptsize

\resizebox{\linewidth}{!}{
\begin{tabular}{@{} l l c l *{7}{c} @{}}
\toprule
Benchmark / Dataset
& Domain
& Input Modality
& Scale
& Arch
& Tooth
& Surf
& MV
& R/CoT
& Closed
& QC \\
\midrule

\rowcolor[gray]{0.95}
\multicolumn{11}{l}{
\textit{\textbf{General Medical VQA Benchmarks}}
} \\

PathVQA~\citep{he2020pathvqa}
\textit{\textsubscript{arXiv'20}}
& Pathology
& 2D Pathology Images
& 4.9K imgs / 32.7K VQA
& \ding{55}
& \ding{55}
& \ding{55}
& \ding{55}
& \ding{55}
& \ding{51}
& \ding{55} \\

OmniMedVQA~\citep{hu2024omnimedvqa}
\textit{\textsubscript{CVPR'24}}
& General
& 2D Medical Images
& 118.0K imgs / 127.9K VQA
& \ding{55}
& \ding{55}
& \ding{55}
& \ding{55}
& \ding{55}
& \ding{51}
& \ding{55} \\

\midrule

\rowcolor[gray]{0.95}
\multicolumn{11}{l}{
\textit{\textbf{Traditional 2D Dental Pathology \& Caries Datasets}}
} \\

DENTEX~\citep{hamamci2023dentex}
\textit{\textsubscript{MICCAI'23}}
& Dental
& 2D Panoramic X-ray
& 2.3K labeled / 1.5K unlabeled imgs
& \ding{55}
& \ding{51}
& \ding{55}
& \ding{55}
& \ding{55}
& \ding{55}
& \ding{51} \\

CariesDet~\citep{tan2025caries}
\textit{\textsubscript{Caries Res.'25}}
& Dental (Caries)
& 2D Bitewing X-ray
& 1.0K imgs / 4 classes
& \ding{55}
& \ding{55}
& \ding{55}
& \ding{55}
& \ding{55}
& \ding{55}
& \ding{51} \\

AlphaDent~\citep{sosnin2025alphadent}
\textit{\textsubscript{arXiv'25}}
& Dental (Caries)
& 2D Intraoral Photo
& 1.3K imgs / 9 classes
& \ding{55}
& \ding{55}
& \ding{55}
& \ding{55}
& \ding{55}
& \ding{55}
& \ding{51} \\

DentalAI~\citep{Dentalai}
\textit{\textsubscript{Kaggle'23}}
& Dental (Caries)
& 2D Intraoral Photo
& 2.4K imgs / 28.9K instances
& \ding{55}
& \ding{55}
& \ding{55}
& \ding{55}
& \ding{55}
& \ding{55}
& \ding{55} \\

\midrule

\rowcolor[gray]{0.95}
\multicolumn{11}{l}{
\textit{\textbf{3D Dental Anatomy \& Segmentation Benchmarks}}
} \\

3DTeethSeg~\citep{ben20233dteethseg}
\textit{\textsubscript{MICCAI'22}}
& Dental
& Native 3D IOS
& 1.8K scans / 23.9K teeth
& \ding{55}
& \ding{51}
& \ding{55}
& \ding{55}
& \ding{55}
& \ding{55}
& \ding{51} \\

ToothFairy2~\citep{bolelli2025segmenting}
\textit{\textsubscript{CVPR'25}}
& Dental
& 3D CBCT
& 530 vols / 42 classes
& \ding{55}
& \ding{51}
& \ding{55}
& \ding{55}
& \ding{55}
& \ding{55}
& \ding{51} \\

\midrule

\rowcolor[gray]{0.95}
\multicolumn{11}{l}{
\textit{\textbf{Specialized Dental VLM Resources \& VQA Benchmarks}}
} \\

MetaDent~\citep{li2026metadent}
\textit{\textsubscript{JDR'26}}
& Dental
& 2D Dental Images
& 60.6K imgs / 18.4K VQA
& \ding{55}
& \ding{55}
& \ding{55}
& \ding{55}
& \ding{55}
& \ding{51}
& \ding{51} \\

MMOral-OPG~\citep{hao2026towards}
\textit{\textsubscript{NeurIPS'25}}
& Dental
& 2D Panoramic X-ray
& 20.5K imgs / 1.3M instr. / 1.0K eval. VQA
& \ding{51}
& \ding{51}
& \ding{55}
& \ding{55}
& \ding{55}
& \ding{51}
& \ding{51} \\

MMOral-Uni~\citep{hao2026oralgpt}
\textit{\textsubscript{CVPR'26}}
& Dental
& 5 Dental Modalities
& 2.8K open VQA
& \ding{55}
& \ding{51}
& \ding{55}
& \ding{55}
& \ding{55}
& \ding{55}
& \ding{51} \\

DentVLM~\citep{meng2026multimodal}
\textit{\textsubscript{Nat. Commun.'26}}
& Dental
& 7 Dental Image Modalities
& 110.4K imgs / 2.4M VQA
& \ding{51}
& \ding{51}
& \ding{55}
& \ding{55}
& \ding{51}
& \ding{51}
& \ding{51} \\

IOSVLM~\citep{xiong2026iosvlm}
\textit{\textsubscript{arXiv'26}}
& Dental
& Native 3D IOS
& 19.0K scans / 249.0K VQA
& \ding{51}
& \ding{55}
& \ding{55}
& \ding{55}
& \ding{55}
& \ding{51}
& \ding{51} \\

\midrule

\rowcolor{gray!15}

\textbf{CariesBench (Ours)}
& \textbf{Dental (Caries)}
& \textbf{3D IOS $\rightarrow$ MV 2D}
& \textbf{5.9K scans $\rightarrow$ 85.4K imgs / 133.9K VQA}
& \textbf{\ding{51}}
& \textbf{\ding{51}}
& \textbf{\ding{51}}
& \textbf{\ding{51}}
& \textbf{\ding{51}}
& \textbf{\ding{51}}
& \textbf{\ding{51}} \\

\bottomrule
\end{tabular}
}


\begin{minipage}{\linewidth}
\scriptsize
\textbf{Criteria:}
\textbf{Arch/Tooth/Surf} denote explicit tasks or evaluation at the corresponding anatomical granularity;
\textbf{MV} denotes complementary views jointly used within one task instance;
\textbf{R/CoT} denotes explicit diagnostic rationales or reasoning traces;
\textbf{Closed} denotes closed-ended VQA or structured fixed-choice questions;
\textbf{QC} denotes documented expert annotation or verification.
\end{minipage}

\end{table*}

\subsection{General Medical Multimodal Benchmarks}
General medical multimodal benchmarks have established visual question answering as a widely used paradigm for evaluating medical image understanding. PathVQA~\citep{he2020pathvqa} focuses on pathology images with open-ended questions grounded in pathological findings, whereas OmniMedVQA~\citep{hu2024omnimedvqa} broadens evaluation across diverse imaging modalities and anatomical regions. These benchmarks provide important testbeds for medical visual understanding, but their evaluation is primarily organized around individual image--question pairs rather than anatomically linked multi-view observations. They therefore do not explicitly capture the hierarchical anatomical structure or cross-view evidence dependencies required for dental diagnosis, in which assessment may progress from the entire dentition to individual teeth and specific tooth surfaces.

\subsection{Dental Multimodal Benchmarks}
Dental benchmarks have progressively evolved from 2D pathology recognition and localization to 3D anatomical modeling and vision-language reasoning. Traditional datasets such as DENTEX~\citep{hamamci2023dentex}, CariesDet~\citep{tan2025caries}, AlphaDent~\citep{sosnin2025alphadent}, and DentalAI~\citep{Dentalai} primarily support recognition, localization, or segmentation of dental abnormalities from panoramic radiographs, bitewing X-rays, or intraoral photographs, while Caries-DETR~\citep{liu2026cariesdetrtoothstructureaware} further investigates DETR-based caries lesion detection in 2D RGB intraoral photographs. In parallel, 3DTeethSeg~\citep{ben20233dteethseg} and ToothFairy2~\citep{bolelli2025segmenting} introduce native 3D IOS or CBCT data, but primarily target anatomical segmentation rather than clinical vision-language reasoning.

More recent dental vision-language benchmarks extend evaluation toward clinical question answering and increasingly diverse multimodal reasoning settings. MetaDent~\citep{li2026metadent} introduces dental VQA over 2D dental images; MMOral-OPG~\citep{hao2026towards} focuses on panoramic X-ray understanding; MMOral-Uni~\citep{hao2026oralgpt} further extends evaluation across five dental imaging modalities and five task types; DentVLM~\citep{meng2026multimodal} broadens dental reasoning across seven imaging modalities and 36 clinical tasks; and IOSVLM~\citep{xiong2026iosvlm}, the closest setting to ours, investigates generative VQA directly over native 3D IOS. However, as summarized in Table~\ref{tab:benchmark_comparison}, these benchmarks cover complementary aspects of dental multimodal understanding, but none of the compared resources jointly evaluates arch-, tooth-, and surface-level reasoning, multi-view evidence integration, and surface-level caries assessment.

CariesBench targets this missing intersection by combining hierarchical anatomical evaluation with multi-view observations derived from the same underlying 3D anatomy. This design enables systematic assessment of anatomical correspondence, cross-view evidence integration, and clinically consistent decisions across arch, tooth, and surface granularities.

\section{Construction of CariesBench}
\label{sec:construction}

\subsection{Benchmark Overview}
Figure~\ref{fig:overview} summarizes the construction of CariesBench, which converts 3D IOS into standardized arch- and tooth-level multi-view 2D representations for existing MLLMs.

Based on these representations, CariesBench organizes evaluation into three complementary task families: Arch-level Clinical Screening (ACS) for global understanding of the dentition, Tooth-level Diagnostic Reasoning (TDR) for fine-grained reasoning over individual teeth and their surfaces, and Structured Clinical Decision-Making (SCD) for objectively scored clinical decisions across anatomical granularities. Together, these task families form a hierarchical evaluation from global arch-level screening to localized tooth- and surface-level reasoning and structured clinical decision-making.

The 5,932 IOS were obtained from 2,966 patients, each contributing one maxillary and one mandibular scan. Using a fixed patient-level split, the 2,966 patients were partitioned into 2,372/297/297 patients for training, validation, and testing, respectively. All IOS and their derived arch-level renderings, tooth-level views, surface observations, and VQA instances from the same patient were confined to the same split before downstream VQA sampling or training-data rebalancing. This patient-disjoint partition prevents patient-level information leakage across training and evaluation sets.

\subsection{3D-to-2D Multi-view Representation}

\textbf{Arch-level Multi-view Rendering.} Each complete maxillary or mandibular IOS is rendered into six standardized views covering complementary anatomical directions. Maxillary scans include right buccal, frontal, left buccal, right palatal, occlusal, and left palatal views. Mandibular scans include right lingual, occlusal, left lingual, right buccal, frontal, and left buccal views. The six views are arranged in a fixed arch-specific \(2\times3\) layout, providing consistent anatomical orientation and complementary observations of the same dental arch.

\textbf{Tooth-level Multi-view Rendering.}
Each complete IOS is first segmented using the pretrained 3DTeethSAM~\citep{lu20263dteethsam} model, and each segmented tooth is rendered into a standardized multi-view panel. Panel 1 provides an overview of the target tooth, followed by anatomically defined surface-specific views. Incisors contain labial, mesial, inner, distal, and incisal views; posterior teeth contain buccal, mesial, inner, distal, and occlusal views; and canines use four corresponding surface-specific views without an additional incisal or occlusal view. Inner surfaces are designated as palatal for maxillary teeth and lingual for mandibular teeth, while anterior and posterior outer surfaces are termed labial and buccal, respectively. Mesial and distal orientations are mirrored between the right and left quadrants to preserve tooth-relative anatomical orientation.

\textbf{Anatomical Correspondence.}
Unlike independently acquired 2D dental images, all views associated with a given arch or tooth are rendered from the same underlying 3D anatomical instance. CariesBench therefore preserves explicit anatomical correspondence across views, enabling evaluation of whether an MLLM can associate findings with the correct tooth and surface, integrate complementary evidence, and maintain a consistent diagnosis across viewpoints.

\subsection{Modified ICDAS-II-Based Clinical Status Taxonomy}

Based on the ICDAS-II caries coding scheme, we define a task-oriented clinical status taxonomy for multi-view IOS assessment. ICDAS-II codes 1 and 2 are merged into a unified early-caries category to reduce ambiguity between subtle early-stage manifestations in rendered IOS views, while the remaining caries categories follow the ordinal progression of ICDAS-II codes 3–6. Because restoration-related conditions are also clinically relevant to dental screening, we additionally introduce two categories representing restored teeth without and with secondary caries. The resulting taxonomy contains eight clinical states, $\mathcal{L}=$\{L0,L1,L2,L3,L4,L5,L6,L7\}. Specifically, L0–L5 denote Sound, Early Caries, Localized Enamel Breakdown, Underlying Dark Shadow, Distinct Cavity, and Extensive Cavity, corresponding to ICDAS-II codes 0, 1–2, 3, 4, 5, and 6, respectively; L6–L7 represent restoration without and with secondary caries and are categorical extensions rather than ordinal severity levels. This eight-state clinical status taxonomy serves as the standardized clinical label space for benchmark construction and evaluation.

\subsection{Hierarchical Clinical Task Construction}

\textbf{Arch-level Clinical Screening (ACS).}
ACS evaluates global understanding of a complete dental arch through six task types: Global Screening (GS), Tooth Enumeration (TE), Specific Region (SR), Differential Diagnosis (DD), Severity Comparison (SC), and Sound Tooth Identification (SI). 

\textbf{Tooth-level Diagnostic Reasoning (TDR).}
TDR evaluates fine-grained reasoning over an individual tooth through Surface Analysis (SA), Full Tooth Scan (FS), and Sound Tooth Verification (SV). These tasks require models to associate observations with the correct anatomical surfaces, integrate complementary views of the same tooth, and derive a clinically consistent diagnosis. For reasoning-oriented instances, the diagnostic process follows a five-step progression: View Identification (VI), Morphology Observation (MO), Texture Details (TD), Cross-view Consistency (CC), and Final Diagnosis (FD). 

\textbf{Structured Clinical Decision-Making (SCD).}
SCD complements open-ended reasoning with objectively scored true/false (TF) and multiple-choice (MC) questions spanning surface, tooth, and arch granularities. These tasks evaluate caries and restoration verification, severity classification, and identification of clinically relevant surfaces or teeth. 

\subsection{VQA Generation and Quality Control}
\label{sec:vqa_genneration}

We construct CariesBench using a structured-annotation-grounded, semi-automated pipeline that explicitly separates structured clinical annotations from subsequent language realization. The structured clinical annotations were derived from epidemiological surveys, oral examinations, and clinical data curation at a university-affiliated stomatological hospital. Annotation and data entry were conducted by five postgraduate dental trainees, three standardized-residency trainees, and one attending dentist under senior clinical supervision. The resulting annotations were organized by patient, tooth, anatomical surface, and clinical status and served as the reference clinical labels for benchmark construction and evaluation. For the open-ended ACS and TDR tasks, GPT-4o was used only for language realization of annotation-grounded questions and responses. Gemini-2.5-Pro subsequently audited each generated instance for consistency with the structured annotations, anatomical correspondence, and unsupported clinical claims. For SCD, questions and target answers were instantiated directly from the structured annotations rather than generated by an LLM. Across all task families, evaluation targets are derived from the structured clinical annotations rather than from LLM-generated reference prose.

Expert validation was further performed on a stratified subset comprising 20\% of the full benchmark, covering all clinical-status categories and task families. Two dentists with 8 and 3 years of clinical experience, respectively, reviewed the selected instances using a standardized five-point clinical evaluation rubric under a protocol designed and supervised by a senior dentist with 15 years of clinical experience. Overall response quality was assessed along five dimensions: Tooth/Surface Mapping, Evidence
Grounding, Target Completeness, Cross-view Consistency, and Conclusion Support. For TDR, the five-step diagnostic explanations were additionally assessed in terms of VI, MO, TD, CC, and FD. Across the reported evaluation criteria, mean expert ratings ranged from 4.26 to 4.95 on the five-point scale. Of the expert-reviewed subset, 800 stratified instances spanning all task families and clinical-status categories were independently assessed by both dentists for inter-rater reliability analysis. Inter-rater agreement was quantified using linearly weighted Cohen's $\kappa$, yielding $\kappa=0.86$. Disagreements were resolved through adjudication and consensus under senior clinical supervision; 5.2\% of reviewed instances required adjudication, while 1.8\% were excluded because of unsupported visual claims or unresolved clinical ambiguity.

\subsection{Benchmark Statistics}

\begin{figure*}[ht]
    \centering
    \includegraphics[width=1\linewidth]{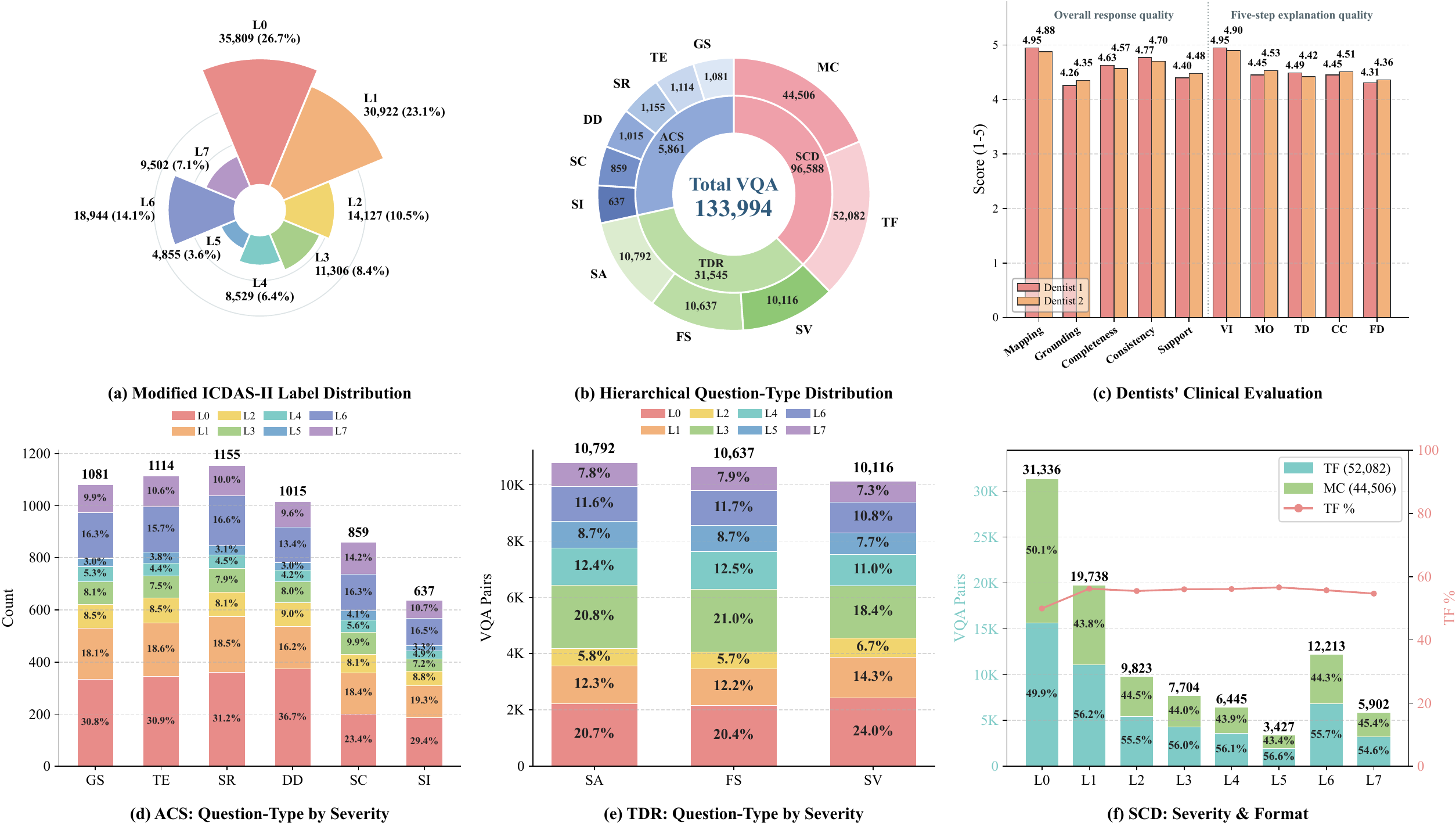}
    \caption{Statistical composition and expert quality assessment of CariesBench.}
    \label{fig:benchmark_statistics}
\end{figure*}

Figure~\ref{fig:benchmark_statistics} summarizes the benchmark statistics and expert quality assessment. The benchmark contains 133,994 VQA instances, including 5,861 ACS, 31,545 TDR, and 96,588 SCD instances, with detailed distributions across their respective question types. It covers all eight modified ICDAS-II clinical states and spans complementary screening, diagnostic reasoning, and structured decision formats. Expert evaluation by two dentists yielded consistently high scores for both generated responses and five-step TDR explanations.


\section{Experiments}
\label{sec:experiments}

\subsection{Experimental Setups}
\textbf{Benchmarked MLLMs.}
 We evaluate 24 representative multimodal large language models (MLLMs) grouped into three categories: commercial, open-source, and medical-specialized models. Commercial models include GPT-5.5~\citep{openai_gpt55}, GPT-5.4-Nano~\citep{openai_gpt54nano}, GLM-5.2~\citep{zeng2026glm}, Claude-Sonnet-5~\citep{anthropic_claude_sonnet_5}, Claude-Haiku-4.5~\citep{anthropic_claude_haiku_4_5}, Gemini-3.6-Flash~\citep{google_gemini_3_6_flash}, and Gemini-3.1-Pro-Preview~\citep{google_gemini_3_1_pro_preview}. Open-source models include LLaVA-1.5-7B~\citep{liu2024improved}, ShareGPT4V-7B~\citep{chen2024sharegpt4v}, InternVL2-8B~\citep{chen2024internvl}, InternVL3.5-8B~\citep{wang2025internvl3}, Qwen2.5-VL-7B-Instruct~\citep{bai2025qwen25vltechnicalreport}, Qwen3-VL-8B-Instruct~\citep{bai2025qwen3}, Qwen3.5-9B~\citep{qwen3.5}, DeepSeek-VL-7B~\citep{lu2024deepseek}, Kimi-VL-A3B-Thinking~\citep{team2025kimi}, and MiMo-VL-7B-RL~\citep{yue2025mimo}. Medical-specialized models include HuatuoGPT-Vision-7B~\citep{zhang2023huatuogpt}, LLaVA-Med-7B~\citep{li2023llava}, DentalGPT-Mid~\citep{cai2025dentalgpt}, HealthGPT-Pro-8B~\citep{lin2025healthgpt}, Lingshu-7B~\citep{xu2026lingshu}, OralGPT-Omni-7B~\citep{hao2026oralgpt}, and DentVLM~\citep{meng2026multimodal}.

 \textbf{Supervised Fine-Tuning.}
We further fine-tune Qwen3-VL-8B-Instruct on the CariesBench training set using LoRA implemented with LLaMA-Factory~\citep{zheng2024llamafactory} for two epochs, with a learning rate of \(2\times10^{-5}\), rank \(r=16\), scaling factor \(\alpha=32\), and BF16 precision. We refer to the resulting model as IntraoralGPT.
 
\textbf{Evaluation Metrics.} Consistent with the hierarchical structure of CariesBench, our primary metric is Clinical Accuracy, which evaluates structured diagnostic correctness after task-specific response parsing, reducing sensitivity to linguistic variation. For ACS, model responses are parsed according to the expected output structure of each question type, with clinical predictions compared against the corresponding structured reference annotations. For TDR, Clinical Accuracy is computed at the task-appropriate anatomical granularity: exact clinical-status matching for the queried surface in SA, mean surface-level exact-match accuracy across the target tooth in FS, and binary sound/non-sound correctness in SV. For SCD, true/false responses are evaluated by binary correctness, whereas multiple-choice responses are evaluated by exact class match. Each task-family Overall is the macro-average of its question types, and Avg. is the macro-average across all 19 question types. 

\textbf{Implementation Details.} For all benchmarked baseline models, we use zero-shot prompting with deterministic decoding (temperature \(=0.0\), top-\(p=1.0\), and repetition penalty \(=1.0\)). Commercial models are evaluated through their official APIs using equivalent decoding settings where supported. All models receive the same task-specific instructions and corresponding rendered inputs. Open-source and medical-specialized models are evaluated locally using PyTorch 2.4.1, Transformers 5.4.0, and CUDA 11.8 with FP16 inference on a server equipped with eight NVIDIA A40 46 GB GPUs.

\subsection{Evaluation Results}
\begin{table*}[ht]
    \centering
    \caption{
Evaluation of representative MLLMs on CariesBench across ACS, TDR, SCD-TF, and SCD-MC. Clinical accuracy (\%) is reported for all 19 sub-tasks; each Overall is the macro-average over its constituent sub-tasks, and Avg. is the macro-average over all 19 sub-tasks. Best and second-best results across all evaluated models are shown in \textbf{bold} and \underline{underlined}, respectively. IntraoralGPT is reported separately as a CariesBench-supervised model.
}
    \label{tab:combined_qtype}
    \renewcommand{\arraystretch}{1.15}
    \resizebox{\linewidth}{!}{
    \begin{tabular}{c l *{24}{c}}
        \toprule
        & & \multicolumn{7}{c}{\textbf{ACS}}
        & \multicolumn{4}{c}{\textbf{TDR}}
        & \multicolumn{6}{c}{\textbf{SCD-TF}}
        & \multicolumn{6}{c}{\textbf{SCD-MC}}
        & \multirow{2}{*}{\textbf{Avg.}} \\
        \cmidrule(lr){3-9}
        \cmidrule(lr){10-13}
        \cmidrule(lr){14-19}
        \cmidrule(lr){20-25}
        & \textbf{Model}
        & GS & TE & SR & DD & SC & SI & \textbf{Overall}
        & SA & FS & SV & \textbf{Overall}
        & Surf & Whl & NoC & Rst & Sec & \textbf{Overall}
        & Wor & SC & Sev & WS & TS & \textbf{Overall}
        & \\
        \midrule

        \multirow{7}{*}{\rotatebox{90}{\textbf{\textit{Commercial}}}}
        & GPT-5.5~\citep{openai_gpt55}
        & 67.9 & 65.9 & 67.0 & 69.6 & 55.9 & 76.9 & 67.2
        & \textbf{31.8} & \textbf{30.5} & \textbf{75.4} & \textbf{45.9}
        & \underline{63.8} & \underline{61.2} & \underline{70.4} & 79.1 & 86.5 & \underline{72.2}
        & \underline{58.7} & \underline{57.4} & \underline{63.8} & 61.5 & 65.2 & \underline{61.3}
        & \underline{63.6} \\

        & GPT-5.4-Nano~\citep{openai_gpt54nano}
        & 70.5 & 66.6 & 67.2 & 69.3 & 56.5 & 75.6 & 67.6
        & 21.8 & 22.4 & 67.2 & 37.1
        & 50.8 & 48.9 & 60.2 & 72.4 & 78.6 & 62.2
        & 45.2 & 43.7 & 49.6 & 46.8 & 52.1 & 47.5
        & 56.1 \\

        & GLM-5.2~\citep{zeng2026glm}
        & 69.6 & \underline{69.5} & 67.7 & 63.8 & 53.2 & 76.3 & 66.7
        & 24.7 & 23.9 & 68.6 & 39.1
        & 51.5 & 49.8 & 62.1 & 73.8 & 79.0 & 63.2
        & 46.3 & 45.1 & 50.8 & 48.0 & 53.5 & 48.7
        & 56.7 \\

        & Claude-Sonnet-5~\citep{anthropic_claude_sonnet_5}
        & 72.3 & 67.6 & 67.8 & 70.1 & 55.8 & 76.8 & 68.4
        & 30.1 & 29.5 & 73.8 & 44.5
        & 61.9 & 59.7 & 68.9 & 77.8 & 84.2 & 70.5
        & 56.1 & 54.8 & 59.5 & 55.4 & 58.9 & 56.9
        & 62.2 \\

        & Claude-Haiku-4.5~\citep{anthropic_claude_haiku_4_5}
        & 71.7 & 61.1 & 68.4 & 70.5 & 60.2 & 70.2 & 67.0
        & 26.9 & 26.1 & 69.7 & 40.9
        & 55.4 & 53.7 & 64.8 & 74.5 & 80.6 & 65.8
        & 49.7 & 48.5 & 54.9 & 52.6 & 56.7 & 52.5
        & 58.7 \\

        & Gemini-3.6-Flash~~\citep{google_gemini_3_6_flash}
        & 70.5 & 64.4 & 62.2 & 69.1 & 57.0 & 76.6 & 66.6
        & 29.4 & 28.8 & 73.1 & 43.8
        & 60.2 & 58.6 & 67.1 & 77.0 & 83.5 & 69.3
        & 54.8 & 53.2 & 59.9 & 57.7 & 61.4 & 57.4
        & 61.3 \\

        & Gemini-3.1-Pro-Preview~\citep{google_gemini_3_1_pro_preview}
        & 70.2 & 66.6 & 64.1 & 67.4 & 60.2 & \underline{77.2} & 67.6
        & 26.6 & 25.7 & 70.2 & 40.8
        & 58.2 & 56.7 & 65.3 & 81.4 & 82.0 & 68.7
        & 52.9 & 51.5 & 54.8 & 52.2 & 66.7 & 55.6
        & 60.5 \\

        \midrule

        \multirow{10}{*}{\rotatebox{90}{\textbf{\textit{Open-Source}}}}
        & LLaVA-1.5-7B~\citep{liu2024improved}
        & 67.8 & 66.1 & 68.7 & 71.7 & 52.7 & 75.8 & 67.1
        & 5.2 & 25.5 & 71.2 & 34.0
        & 41.1 & 47.0 & 40.6 & 76.7 & 29.8 & 47.0
        & 32.4 & 31.2 & 29.5 & 60.3 & 21.6 & 35.0
        & 48.2 \\

        & ShareGPT4V-7B~\citep{chen2024sharegpt4v}
        & 69.2 & 67.0 & \textbf{70.8} & 70.2 & 57.6 & 75.6 & 68.4
        & 19.5 & 24.9 & 72.5 & 39.0
        & 32.6 & 55.5 & 64.7 & 15.0 & 5.7 & 34.7
        & 38.6 & 36.9 & 31.1 & \underline{67.7} & 13.8 & 37.6
        & 46.8 \\

        & InternVL2-8B~\citep{chen2024internvl}
        & 67.4 & 69.2 & 64.2 & 68.4 & 59.3 & 76.1 & 67.4
        & 19.4 & 22.0 & 71.9 & 37.8
        & 38.3 & 41.7 & 64.7 & 73.1 & 5.8 & 44.7
        & 44.1 & 42.6 & 24.1 & 45.6 & 66.1 & 44.5
        & 50.7 \\

        & InternVL3.5-8B~~\citep{wang2025internvl3}
        & 67.2 & 62.7 & 68.7 & 69.3 & 59.7 & 76.0 & 67.3
        & 24.9 & 25.0 & 72.0 & 40.6
        & 51.7 & 45.0 & 64.7 & 15.2 & 5.7 & 36.5
        & 49.5 & 47.8 & 29.3 & 38.7 & 51.8 & 43.4
        & 48.7 \\

        & Qwen2.5-VL-7B-Instruct~\citep{bai2025qwen25vltechnicalreport}
        & \textbf{72.5} & 59.5 & 69.5 & 70.1 & 61.7 & 76.5 & 68.3
        & 16.2 & 24.0 & 72.1 & 37.4
        & 40.8 & 47.6 & 51.2 & 80.0 & 92.5 & 62.4
        & 39.8 & 37.6 & 15.8 & 52.7 & 4.2 & 30.0
        & 51.8 \\

        & Qwen3-VL-8B-Instruct~\citep{bai2025qwen3}
        & 65.2 & 66.4 & 69.9 & 69.9 & 62.2 & 75.1 & 68.1
        & 24.3 & 23.0 & 55.9 & 34.4
        & 49.1 & 39.3 & 64.7 & 68.6 & 81.9 & 60.7
        & 54.0 & 51.8 & 20.7 & 43.9 & 15.4 & 37.2
        & 52.7 \\

        & Qwen3.5-9B~\citep{qwen3.5}
        & 70.7 & 67.0 & 65.1 & 68.0 & \underline{63.5} & 76.5 & \underline{68.5}
        & 27.8 & \underline{29.8} & 71.8 & 43.1
        & 11.2 & 60.9 & 63.8 & 2.2 & 0.4 & 27.7
        & 52.6 & 50.4 & 4.6 & 3.4 & \underline{74.0} & 37.0
        & 45.5 \\

        & DeepSeek-VL-7B~\citep{lu2024deepseek}
        & 71.7 & 67.5 & 68.4 & 68.6 & 52.6 & 76.7 & 67.6
        & 9.0 & 29.2 & 69.9 & 36.0
        & 60.6 & 51.4 & 64.7 & 19.2 & 5.8 & 40.3
        & 42.8 & 40.6 & 8.7 & 19.7 & 20.0 & 26.4
        & 44.6 \\

        & Kimi-VL-A3B-Thinking~\citep{team2025kimi}
        & 67.0 & 63.6 & 69.5 & 68.8 & 62.2 & 76.2 & 67.9
        & 22.6 & 25.6 & 71.9 & 40.0
        & 60.8 & 46.0 & 64.9 & 70.1 & 82.5 & 64.9
        & 51.4 & 49.8 & 33.7 & 50.1 & 25.1 & 42.0
        & 55.9 \\

        & MiMo-VL-7B-RL~\citep{yue2025mimo}
        & 71.8 & 68.9 & 66.9 & 67.9 & \textbf{65.7} & 76.8 & \textbf{69.7}
        & 28.1 & 29.4 & 73.0 & 43.5
        & 59.6 & 51.2 & 64.7 & 19.4 & 5.8 & 40.1
        & 54.0 & 52.1 & 24.0 & 15.0 & 67.5 & 42.5
        & 50.6 \\

        \midrule

        \multirow{7}{*}{\rotatebox{90}{\textbf{\textit{Medical-Specialized}}}}
        & HuatuoGPT-Vision-7B~\citep{zhang2023huatuogpt}
        & 69.7 & 69.1 & \underline{70.7} & 67.9 & 52.7 & 75.8 & 67.7
        & 12.8 & 27.6 & 70.3 & 36.9
        & 59.3 & 51.4 & 64.7 & 14.9 & 5.7 & 39.2
        & 43.2 & 41.7 & 13.6 & 50.2 & 19.9 & 33.7
        & 46.4 \\

        & LLaVA-Med-7B~\citep{li2023llava}
        & 71.7 & 69.4 & 68.7 & 64.2 & 52.6 & 75.5 & 67.0
        & 16.8 & 18.9 & 66.4 & 34.0
        & 40.8 & 48.6 & 35.3 & \underline{85.1} & \textbf{94.3} & 60.8
        & 28.4 & 27.1 & 0.0 & 0.2 & 0.0 & 11.1
        & 45.5 \\

        & DentalGPT-Mid~\citep{cai2025dentalgpt}
        & 66.9 & 62.7 & 65.3 & 71.5 & 54.9 & 75.9 & 66.2
        & 8.2 & 10.7 & 72.7 & 30.5
        & 49.5 & 48.6 & 63.4 & 81.6 & 31.3 & 54.9
        & 38.6 & 37.2 & 14.1 & 48.2 & 7.8 & 29.2
        & 47.8 \\

        & HealthGPT-Pro-8B~\citep{lin2025healthgpt}
        & 68.1 & 27.5 & 68.7 & 71.3 & 56.5 & 75.7 & 61.3
        & 17.4 & 26.4 & 72.1 & 38.6
        & 55.0 & 45.8 & 64.7 & 78.7 & 60.1 & 60.9
        & 49.3 & 47.8 & 51.5 & 55.4 & 29.8 & 46.8
        & 53.8 \\

        & Lingshu-7B~\citep{xu2026lingshu}
        & 68.3 & 62.6 & 68.1 & 68.3 & 57.0 & 75.8 & 66.7
        & 22.7 & 12.1 & 71.5 & 35.4
        & 58.1 & 51.0 & 64.7 & 24.9 & 31.5 & 46.0
        & 46.8 & 45.2 & 39.6 & 59.8 & 48.8 & 48.0
        & 51.4 \\

        & OralGPT-Omni-7B~\citep{hao2026oralgpt}
        & 66.6 & 69.2 & 61.5 & 71.6 & 57.9 & 72.9 & 66.6
        & 10.5 & 17.4 & 39.5 & 22.5
        & 61.7 & 51.4 & 64.7 & 84.0 & \underline{94.0} & 71.2
        & 55.6 & 53.8 & 12.7 & 52.7 & 21.6 & 39.3
        & 53.6 \\

        & DentVLM~\citep{meng2026multimodal}
        & \underline{72.4} & \textbf{69.7} & 68.7 & \underline{72.1} & 52.6 & 75.0 & 68.4
        & 23.6 & 26.9 & 71.8 & 40.8
        & 50.8 & 48.7 & 52.1 & 47.8 & 93.1 & 58.5
        & 57.9 & 56.3 & 23.2 & 24.6 & 28.5 & 38.1
        & 53.5 \\

        & IntraoralGPT (ours)
        & 70.5 & 68.9 & 68.1 & \textbf{72.4} & 53.2 & \textbf{78.0} & \underline{68.5}
        & \underline{30.6} & 29.1 & \underline{74.5} & \underline{44.7}
        & \textbf{78.1} & \textbf{80.8} & \textbf{72.7} & \textbf{86.1} & \textbf{94.3} & \textbf{82.4}
        & \textbf{70.5} & \textbf{68.9} & \textbf{77.0} & \textbf{73.1} & \textbf{74.2} & \textbf{72.7}
        & \textbf{69.5} \\

        \bottomrule
    \end{tabular}
    }
\end{table*}

Based on Tables~\ref{tab:combined_qtype} and~\ref{tab:sft_hierarchical}, we summarize four key findings on CariesBench.

\textbf{Finding 1. Surface-specific tooth-level reasoning is the most pronounced bottleneck observed on CariesBench.}
The lowest performance is observed on SA and FS, which require fine-grained surface-level clinical-status prediction from tooth-specific visual observations. For example, GPT-5.5 achieves 31.8\% on SA and 30.5\% on FS, compared with 75.4\% on SV. Across all 24 baselines, SA and FS peak at only 31.8\% and 30.5\%, respectively, whereas SV reaches 75.4\%. This contrast suggests that the largest performance degradation is associated with tasks requiring surface-specific localization and cross-view integration rather than tooth-level input alone.

\noindent\textbf{Finding 2. MLLMs exhibit a pronounced benchmark-level global-to-local performance disparity.} Across CariesBench, models perform consistently worse on TDR than on ACS under the respective task-specific evaluation protocols. Across the 24 baselines, the mean ACS Overall accuracy is 67.3\%, compared with 38.2\% for TDR Overall, yielding a descriptive benchmark-level performance difference of 29.1 percentage points. Because ACS and TDR differ in anatomical scope, prediction targets, diagnostic complexity, answer formats, and scoring units, this observed difference cannot identify the specific contribution of anatomical granularity, cross-view integration, or other task-related factors.

\textbf{Finding 3. Dental or medical specialization does not consistently translate into stronger multi-view intraoral reasoning.}
Under the same zero-shot evaluation protocol, medical- and dental-specialized models do not consistently outperform general-purpose MLLMs on CariesBench. DentVLM, for example, achieves competitive results on TE (69.7\%) and FS (26.9\%), but its TDR Overall remains 40.8\%, below several general-purpose commercial models. OralGPT-Omni-7B and DentalGPT-Mid achieve 22.5\% and 30.5\% TDR Overall, respectively, whereas GPT-5.5 achieves the strongest zero-shot baseline TDR performance at 45.9\%. These results suggest that domain specialization alone is insufficient to ensure strong performance on the multi-view reasoning requirements evaluated by CariesBench, and that targeted supervision explicitly representing anatomical correspondence and cross-view evidence may be beneficial.

\begin{table*}[ht]
    \centering
    \caption{
    Effect of CariesBench-specific supervision across the hierarchical clinical evaluation.
    Values denote mean clinical accuracy (\%) over question types within each task group.
    $\Delta$ denotes the absolute improvement of IntraoralGPT over its Qwen3-VL-8B-Instruct base model.
    }
    \label{tab:sft_hierarchical}
    \setlength{\tabcolsep}{5pt}
    \begin{tabular}{lcccc}
        \toprule
        \textbf{Task Group}
        & \textbf{Types}
        & \textbf{Qwen3-VL-8B-Instruct}
        & \textbf{IntraoralGPT}
        & \textbf{$\Delta$} \\
        \midrule
        \textbf{ACS}
        & 6
        & 68.1
        & \textbf{68.5}
        & +0.4 \\

        \textbf{TDR}
        & 3
        & 34.4
        & \textbf{44.7}
        & +10.3 \\

        \textbf{SCD-TF}
        & 5
        & 60.7
        & \textbf{82.4}
        & +21.7 \\

        \textbf{SCD-MC}
        & 5
        & 37.2
        & \textbf{72.7}
        & +35.5 \\

        \midrule
        \textbf{All Tasks}
        & 19
        & 52.7
        & \textbf{69.5}
        & \textbf{+16.8} \\
        \bottomrule
    \end{tabular}
\end{table*}

\textbf{Finding 4. CariesBench-specific supervision substantially improves hierarchical clinical reasoning.}
To investigate whether the limitations exposed by CariesBench can be mitigated through benchmark-specific supervision, we compare IntraoralGPT with its Qwen3-VL-8B-Instruct base model under the same evaluation protocol. IntraoralGPT improves the 19-task macro-average clinical accuracy from 52.7\% to 69.5\%, corresponding to a gain of 16.8 percentage points. The gains vary substantially across task groups: ACS improves by 0.4 points, TDR by 10.3 points, SCD-TF by 21.7 points, and SCD-MC by 35.5 points. Within TDR, SA improves from 24.3\% to 30.6\%, while FS improves from 23.0\% to 29.1\%. These results show that CariesBench-specific supervision yields particularly large gains on structured clinical decisions while also improving tooth-level multi-view reasoning. Nevertheless, IntraoralGPT reaches 44.7\% TDR Overall, remaining slightly below the strongest zero-shot baseline, GPT-5.5, at 45.9\%, highlighting the remaining difficulty of fine-grained anatomical localization and cross-view evidence integration.

\subsection{Case Study}

\begin{figure*}[ht]
    \centering
    \includegraphics[width=1\linewidth]{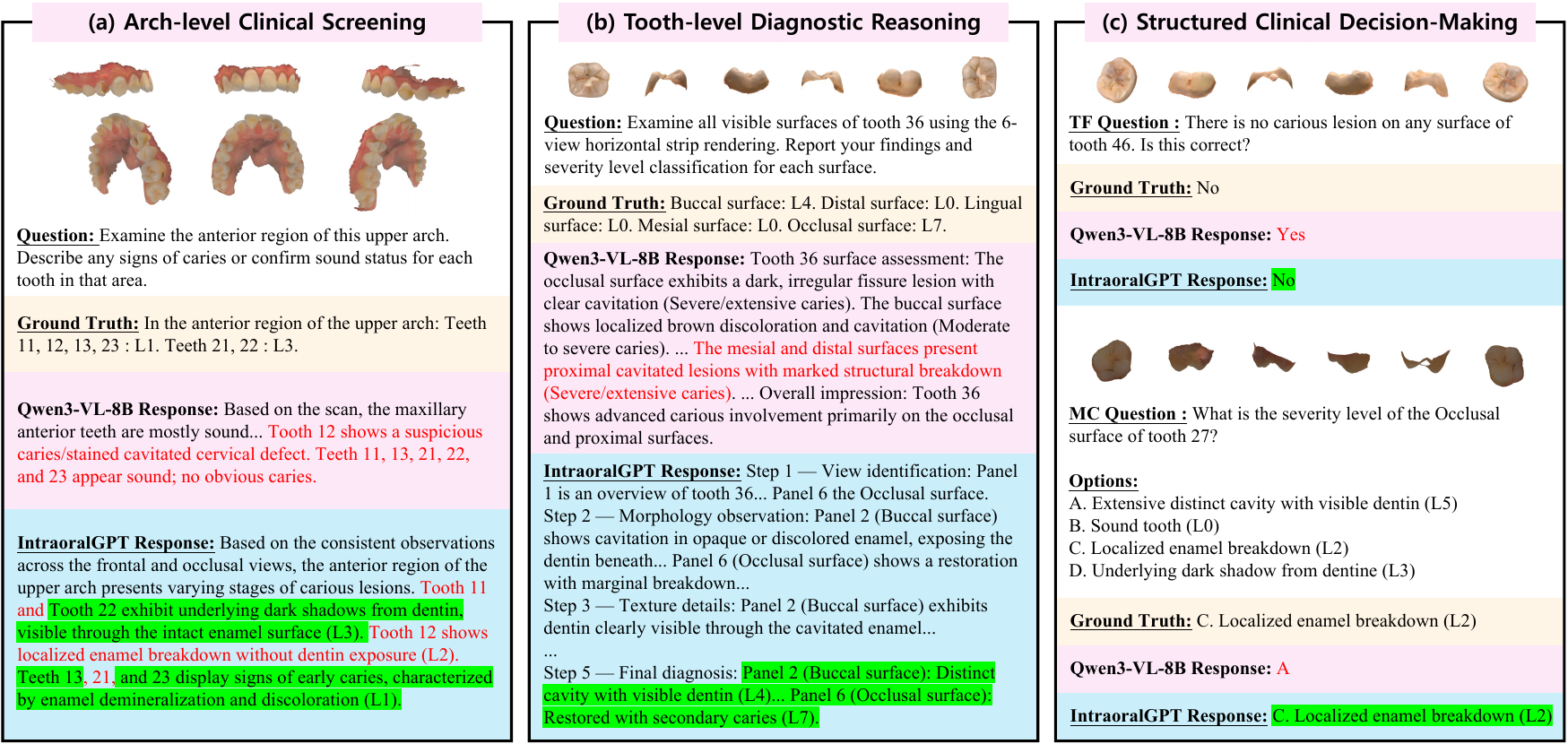}
    \caption{Representative case studies on CariesBench across (a) ACS, (b) TDR, and (c) SCD. Comparisons between Qwen3-VL-8B-Instruct and IntraoralGPT illustrate their differences in hierarchical dental reasoning. Correct predictions are highlighted in green and incorrect predictions in red.}
    \label{fig:case_study}
\end{figure*}

Figure~\ref{fig:case_study} qualitatively compares Qwen3-VL-8B-Instruct with its CariesBench-supervised counterpart, IntraoralGPT, across the three evaluation levels. In the ACS example, IntraoralGPT produces a more accurate tooth-specific clinical assessment, although residual grading errors remain. In the TDR example, IntraoralGPT more accurately associates the observed abnormality with the corresponding tooth surface, consistent with its improved quantitative performance on tooth-level multi-view tasks. In the illustrated SCD examples, IntraoralGPT correctly answers both structured decisions whereas Qwen3-VL-8B-Instruct fails both. These examples qualitatively mirror the task-wise gains in Table~\ref{tab:sft_hierarchical}; consistent with the quantitative results, fine-grained surface-level reasoning nevertheless remains challenging.


\section{Conclusion}

In this work, we introduce CariesBench, a hierarchical benchmark for evaluating clinically grounded multi-view reasoning from 3D intraoral scans in multimodal large language models (MLLMs). By organizing evaluation into Arch-level Clinical Screening (ACS), Tooth-level Diagnostic Reasoning (TDR), and Structured Clinical Decision-Making (SCD), CariesBench evaluates the progression from global arch-level screening to localized tooth- and surface-level reasoning and structured clinical decisions. Extensive evaluation of 24 representative MLLMs reveals a pronounced benchmark-level performance disparity between arch-level screening and tooth-level diagnostic tasks. Controlled multi-view ablations further provide task-dependent evidence for the contribution of complementary visual observations to tooth-level diagnosis. CariesBench-specific supervision substantially improves hierarchical clinical reasoning, increasing the 19-task macro-average clinical accuracy of Qwen3-VL-8B-Instruct from 52.7\% to 69.5\% in IntraoralGPT. Nevertheless, the remaining limitations on TDR highlight the continued difficulty of fine-grained anatomical localization and surface-level clinical-status prediction. CariesBench therefore provides a clinically structured testbed for evaluating MLLMs across anatomical granularities in digital dentistry. Future work should further examine native 3D reasoning and generalization across acquisition settings and clinical populations.

\bibliography{egbib}

@article{he2020pathvqa,
  title={Pathvqa: 30000+ questions for medical visual question answering},
  author={He, Xuehai and Zhang, Yichen and Mou, Luntian and Xing, Eric and Xie, Pengtao},
  journal={arXiv preprint arXiv:2003.10286},
  year={2020}
}

@inproceedings{liu2021slake,
  title={Slake: A semantically-labeled knowledge-enhanced dataset for medical visual question answering},
  author={Liu, Bo and Zhan, Li-Ming and Xu, Li and Ma, Lin and Yang, Yan and Wu, Xiao-Ming},
  booktitle={2021 IEEE 18th international symposium on biomedical imaging (ISBI)},
  pages={1650--1654},
  year={2021},
  organization={IEEE}
}

@inproceedings{hu2024omnimedvqa,
  title={Omnimedvqa: A new large-scale comprehensive evaluation benchmark for medical lvlm},
  author={Hu, Yutao and Li, Tianbin and Lu, Quanfeng and Shao, Wenqi and He, Junjun and Qiao, Yu and Luo, Ping},
  booktitle={2024 IEEE/CVF Conference on Computer Vision and Pattern Recognition (CVPR)},
  pages={22170--22183},
  year={2024},
  organization={IEEE}
}

@article{li2026metadent,
  title={MetaDent: Labeling Clinical Images for Vision-Language Models in Dentistry},
  author={Li, M-X and Deng, W-H and Wu, Z-X and Jin, C-X and Wu, J-M and Han, Yue and Tsoi, James Kit Hon and Xia, G-S and Huang, Cui},
  journal={Journal of Dental Research},
  pages={00220345261424242},
  year={2026},
  publisher={SAGE Publications Sage CA: Los Angeles, CA}
}

@article{hao2026towards,
  title={Towards better dental ai: A multimodal benchmark and instruction dataset for panoramic x-ray analysis},
  author={Hao, Jing and Fan, Yuxuan and Sun, Yanpeng and Guo, Kaixin and Lizhuo, Lin and Yang, Jinrong and Ai, Qiyong and Wong, Lun and Tang, Hao and Hung, Kuo},
  journal={Advances in Neural Information Processing Systems},
  volume={38},
  year={2026}
}

@article{hamamci2023dentex,
  title={Dentex: An abnormal tooth detection with dental enumeration and diagnosis benchmark for panoramic x-rays},
  author={Hamamci, Ibrahim Ethem and Er, Sezgin and Simsar, Enis and Yuksel, Atif Emre and Gultekin, Sadullah and Ozdemir, Serife Damla and Yang, Kaiyuan and Li, Hongwei Bran and Pati, Sarthak and Stadlinger, Bernd and others},
  journal={arXiv preprint arXiv:2305.19112},
  year={2023}
}

@article{meng2026multimodal,
  title={A multimodal vision-language model for comprehensive dental diagnosis and enhanced clinical practice},
  author={Meng, Zijie and Hao, Jin and Dai, Xiwei and Feng, Yang and Liu, Jiaxiang and Feng, Bin and Wu, Huikai and Gai, Xiaotang and Zhu, Hengchuan and Hu, Tianxiang and others},
  journal={Nature Communications},
  year={2026},
  publisher={Nature Publishing Group UK London}
}

@article{xiong2026iosvlm,
  title={IOSVLM: A 3D Vision-Language Model for Unified Dental Diagnosis from Intraoral Scans},
  author={Xiong, Huimin and Meng, Zijie and Hu, Tianxiang and Zhou, Chenyi and Feng, Yang and Liu, Zuozhu},
  journal={arXiv preprint arXiv:2603.16781},
  year={2026}
}

@article{li2023llava,
  title={Llava-med: Training a large language-and-vision assistant for biomedicine in one day},
  author={Li, Chunyuan and Wong, Cliff and Zhang, Sheng and Usuyama, Naoto and Liu, Haotian and Yang, Jianwei and Naumann, Tristan and Poon, Hoifung and Gao, Jianfeng},
  journal={Advances in neural information processing systems},
  volume={36},
  pages={28541--28564},
  year={2023}
}

@article{dikmen2015icdas,
  title={Icdas II criteria (international caries detection and assessment system)},
  author={Dikmen, Benin},
  journal={Journal of Istanbul University Faculty of Dentistry},
  volume={49},
  number={3},
  pages={63--72},
  year={2015},
  publisher={Istanbul University}
}

@article{ntovas2023occlusal,
  title={Occlusal caries detection on 3D models obtained with an intraoral scanner. A validation study},
  author={Ntovas, P and Michou, S and Benetti, AR and Bakhshandeh, A and Ekstrand, K and Rahiotis, C and Kakaboura, A},
  journal={Journal of Dentistry},
  volume={131},
  pages={104457},
  year={2023},
  publisher={Elsevier}
}

@article{eggmann2024recent,
  title={Recent advances in intraoral scanners},
  author={Eggmann, Florin and Blatz, MB},
  journal={Journal of Dental Research},
  volume={103},
  number={13},
  pages={1349--1357},
  year={2024},
  publisher={SAGE Publications Sage CA: Los Angeles, CA}
}

@inproceedings{lu20263dteethsam,
  title={3DTeethSAM: Taming SAM2 for 3D Teeth Segmentation},
  author={Lu, Zhiguo and Lou, Jianwen and Ma, Mingjun and Jin, Hairong and Zheng, Youyi and Zhou, Kun},
  booktitle={Proceedings of the AAAI Conference on Artificial Intelligence},
  volume={40},
  number={9},
  pages={7609--7617},
  year={2026}
}

@article{ismail2007international,
  title={The International Caries Detection and Assessment System (ICDAS): an integrated system for measuring dental caries},
  author={Ismail, Amid I and Sohn, Woosung and Tellez, Marisol and Amaya, Ashley and Sen, Ananda and Hasson, Hana and Pitts, Nigel Berry},
  journal={Community dentistry and oral epidemiology},
  volume={35},
  number={3},
  pages={170--178},
  year={2007},
  publisher={Wiley Online Library}
}

@inproceedings{chen2024sharegpt4v,
  title={Sharegpt4v: Improving large multi-modal models with better captions},
  author={Chen, Lin and Li, Jinsong and Dong, Xiaoyi and Zhang, Pan and He, Conghui and Wang, Jiaqi and Zhao, Feng and Lin, Dahua},
  booktitle={European Conference on Computer Vision},
  pages={370--387},
  year={2024},
  organization={Springer}
}

@inproceedings{liu2024improved,
  title={Improved baselines with visual instruction tuning},
  author={Liu, Haotian and Li, Chunyuan and Li, Yuheng and Lee, Yong Jae},
  booktitle={2024 IEEE/CVF Conference on Computer Vision and Pattern Recognition (CVPR)},
  pages={26286--26296},
  year={2024},
  organization={IEEE}
}

@inproceedings{chen2024internvl,
  title={Internvl: Scaling up vision foundation models and aligning for generic visual-linguistic tasks},
  author={Chen, Zhe and Wu, Jiannan and Wang, Wenhai and Su, Weijie and Chen, Guo and Xing, Sen and Zhong, Muyan and Zhang, Qinglong and Zhu, Xizhou and Lu, Lewei and others},
  booktitle={Proceedings of the IEEE/CVF conference on computer vision and pattern recognition},
  pages={24185--24198},
  year={2024}
}

@article{bai2025qwen3,
  title={Qwen3-vl technical report},
  author={Bai, Shuai and Cai, Yuxuan and Chen, Ruizhe and Chen, Keqin and Chen, Xionghui and Cheng, Zesen and Deng, Lianghao and Ding, Wei and Gao, Chang and Ge, Chunjiang and others},
  journal={arXiv preprint arXiv:2511.21631},
  year={2025}
}

@misc{bai2025qwen25vltechnicalreport,
      title={Qwen2.5-VL Technical Report}, 
      author={Shuai Bai and Keqin Chen and Xuejing Liu and Jialin Wang and Wenbin Ge and Sibo Song and Kai Dang and Peng Wang and Shijie Wang and Jun Tang and Humen Zhong and Yuanzhi Zhu and Mingkun Yang and Zhaohai Li and Jianqiang Wan and Pengfei Wang and Wei Ding and Zheren Fu and Yiheng Xu and Jiabo Ye and Xi Zhang and Tianbao Xie and Zesen Cheng and Hang Zhang and Zhibo Yang and Haiyang Xu and Junyang Lin},
      year={2025},
      eprint={2502.13923},
      archivePrefix={arXiv},
      primaryClass={cs.CV},
      url={https://arxiv.org/abs/2502.13923}, 
}

@inproceedings{zhang2023huatuogpt,
  title={Huatuogpt, towards taming language model to be a doctor},
  author={Zhang, Hongbo and Chen, Junying and Jiang, Feng and Yu, Fei and Chen, Zhihong and Chen, Guiming and Li, Jianquan and Wu, Xiangbo and Zhiyi, Zhang and Xiao, Qingying and others},
  booktitle={Findings of the association for computational linguistics: EMNLP 2023},
  pages={10859--10885},
  year={2023}
}

@article{xu2026lingshu,
  title={Lingshu: Generalist Foundation Model for Unified Multimodal Medical Understanding and Reasoning},
  author={Xu, Weiwen and Chan, Hou Pong and Li, Long and Aljunied, Mahani and Yuan, Ruifeng and Wang, Jianyu and Xiao, Chenghao and Chen, Guizhen and Liu, Chaoqun and Li, Zhaodonghui and others},
  journal={IEEE Transactions on Pattern Analysis and Machine Intelligence},
  year={2026},
  publisher={IEEE}
}

@inproceedings{hao2026oralgpt,
  title={Oralgpt-omni: A versatile dental multimodal large language model},
  author={Hao, Jing and Liang, Yuci and Lin, Lizhuo and Fan, Yuxuan and Zhou, Wenkai and Guo, Kaixin and Ye, Zanting and Sun, Yanpeng and Zhang, Xinyu and Yang, Yanqi and others},
  booktitle={Proceedings of the IEEE/CVF Conference on Computer Vision and Pattern Recognition},
  pages={38509--38519},
  year={2026}
}

@article{cai2025dentalgpt,
  title={Dentalgpt: Incentivizing multimodal complex reasoning in dentistry},
  author={Cai, Zhenyang and Zhang, Jiaming and Zhao, Junjie and Zeng, Ziyi and Li, Yanchao and Liang, Jingyi and Chen, Junying and Yang, Yunjin and You, Jiajun and Deng, Shuzhi and others},
  journal={arXiv preprint arXiv:2512.11558},
  year={2025}
}

@article{zeng2026glm,
  title={Glm-5: from vibe coding to agentic engineering},
  author={Zeng, Aohan and Lv, Xin and Hou, Zhenyu and Du, Zhengxiao and Zheng, Qinkai and Chen, Bin and Yin, Da and Ge, Chendi and Huang, Chenghua and Xie, Chengxing and others},
  journal={arXiv preprint arXiv:2602.15763},
  year={2026}
}

@misc{openai_gpt55,
  author = {{OpenAI}},
  title = {Models API Reference - {GPT-5.5}},
  howpublished = {\url{https://developers.openai.com/api/docs/models/gpt-5.5}},
  year = {2026},
  note = {Accessed: 2026-09-08}
}

@misc{openai_gpt54nano,
  author = {{OpenAI}},
  title = {Models API Reference - {GPT-5.4-nano}},
  howpublished = {\url{https://developers.openai.com/api/docs/models/gpt-5.4-nano}},
  year = {2026},
  note = {Accessed: 2026-09-08}
}

@misc{anthropic_claude_sonnet_5,
  author = {{Anthropic}},
  title = {Introducing {Claude Sonnet 5}},
  howpublished = {\url{https://www.anthropic.com/news/claude-sonnet-5}},
  year = {2026},
  note = {Accessed: 2026-09-08}
}

@misc{anthropic_claude_haiku_4_5,
  author = {{Anthropic}},
  title = {Introducing {Claude Haiku 4.5}},
  howpublished = {\url{https://www.anthropic.com/news/claude-haiku-4-5}},
  year = {2025},
  note = {Accessed: 2026-09-08}
}

@misc{google_gemini_3_1_pro_preview,
  author = {{Google}},
  title = {Gemini API Documentation: {Gemini 3.1 Pro Preview}},
  howpublished = {\url{https://ai.google.dev/gemini-api/docs/models/gemini-3.1-pro-preview}},
  year = {2026},
  note = {Accessed: 2026-09-08}
}

@misc{google_gemini_3_6_flash,
  author = {{Google}},
  title = {Gemini API Documentation: {Gemini 3.6 Flash}},
  howpublished = {\url{https://ai.google.dev/gemini-api/docs/models/gemini-3.6-flash}},
  year = {2026},
  note = {Accessed: 2026-09-08}
}

@article{sosnin2025alphadent,
  title={AlphaDent: A dataset for automated tooth pathology detection},
  author={Sosnin, Evgeniy I and Vasilev, Yuriy L and Solovyev, Roman A and Stempkovskiy, Aleksandr L and Telpukhov, Dmitry V and Vasilev, Artem A and Amerikanov, Aleksandr A and Romanov, Aleksandr Y},
  journal={arXiv preprint arXiv:2507.22512},
  year={2025}
}

@dataset{Dentalai,
  author={Pawan Valluri},
  title={Dentalai Computer Vision Project},
  year={2023},
  url={https://www.kaggle.com/datasets/pawanvalluri/dental-segmentation}
}

@article{ben20233dteethseg,
  title={3DTeethSeg'22: 3D teeth scan segmentation and labeling challenge},
  author={Ben-Hamadou, Achraf and Smaoui, Oussama and Rekik, Ahmed and Pujades, Sergi and Boyer, Edmond and Lim, Hoyeon and Kim, Minchang and Lee, Minkyung and Chung, Minyoung and Shin, Yeong-Gil and others},
  journal={arXiv preprint arXiv:2305.18277},
  year={2023}
}

@inproceedings{bolelli2025segmenting,
  title={Segmenting maxillofacial structures in CBCT volumes},
  author={Bolelli, Federico and Marchesini, Kevin and Van Nistelrooij, Niels and Lumetti, Luca and Pipoli, Vittorio and Ficarra, Elisa and Vinayahalingam, Shankeeth and Grana, Costantino},
  booktitle={2025 IEEE/CVF Conference on Computer Vision and Pattern Recognition (CVPR)},
  pages={5238--5248},
  year={2025},
  organization={IEEE}
}

@article{tan2025caries,
  title={Caries detection in primary molars with bitewing radiographs through deep learning based-object detectors},
  author={Tan, Zhi Qin and Pinho, Ilana Felix and Banks, Ryan and Chen, Xiongjie and Pontes, Laura Regina Antunes and Braga, Mariana Minatel and Mendes, Fausto Medeiros and Li, Yunpeng},
  journal={Caries research},
  year={2025}
}

@article{wang2025internvl3,
  title={Internvl3. 5: Advancing open-source multimodal models in versatility, reasoning, and efficiency},
  author={Wang, Weiyun and Gao, Zhangwei and Gu, Lixin and Pu, Hengjun and Cui, Long and Wei, Xingguang and Liu, Zhaoyang and Jing, Linglin and Ye, Shenglong and Shao, Jie and others},
  journal={arXiv preprint arXiv:2508.18265},
  year={2025}
}

@misc{qwen3.5, title = {{Qwen3.5}: Towards Native Multimodal Agents}, author = {{Qwen Team}}, month = {February}, year = {2026}, url = {https://qwen.ai/blog?id=qwen3.5} }

@article{lin2025healthgpt,
  title={Healthgpt: A medical large vision-language model for unifying comprehension and generation via heterogeneous knowledge adaptation},
  author={Lin, Tianwei and Zhang, Wenqiao and Li, Sijing and Yuan, Yuqian and Yu, Binhe and Li, Haoyuan and He, Wanggui and Jiang, Hao and Li, Mengze and Song, Xiaohui and others},
  journal={arXiv preprint arXiv:2502.09838},
  year={2025}
}

@article{lu2024deepseek,
  title={Deepseek-vl: towards real-world vision-language understanding},
  author={Lu, Haoyu and Liu, Wen and Zhang, Bo and Wang, Bingxuan and Dong, Kai and Liu, Bo and Sun, Jingxiang and Ren, Tongzheng and Li, Zhuoshu and Yang, Hao and others},
  journal={arXiv preprint arXiv:2403.05525},
  year={2024}
}

@article{team2025kimi,
  title={Kimi-vl technical report},
  author={Team, Kimi and Du, Angang and Yin, Bohong and Xing, Bowei and Qu, Bowen and Wang, Bowen and Chen, Cheng and Zhang, Chenlin and Du, Chenzhuang and Wei, Chu and others},
  journal={arXiv preprint arXiv:2504.07491},
  year={2025}
}

@article{yue2025mimo,
  title={Mimo-vl technical report},
  author={Yue, Zihao and Lin, Zhenru and Song, Yifan and Wang, Weikun and Ren, Shuhuai and Gu, Shuhao and Li, Shicheng and Li, Peidian and Zhao, Liang and Li, Lei and others},
  journal={arXiv preprint arXiv:2506.03569},
  year={2025}
}

@inproceedings{zheng2024llamafactory,
  title={Llamafactory: Unified efficient fine-tuning of 100+ language models},
  author={Zheng, Yaowei and Zhang, Richong and Zhang, Junhao and Ye, Yanhan and Luo, Zheyan},
  booktitle={Proceedings of the 62nd annual meeting of the association for computational linguistics (volume 3: system demonstrations)},
  pages={400--410},
  year={2024}
}

@misc{liu2026cariesdetrtoothstructureaware,
      title={Caries DETR: Tooth Structure-aware Prior and Lesion-aware Dynamic Loss Refinement for DETR Based Caries Detection}, 
      author={Xuefen Liu and Xinquan Yang and Mianjie Zheng and Kun Tang and Xuguang Li and Xiaoqi Guo and Linlin Shen and He Meng},
      year={2026},
      eprint={2604.23718},
      archivePrefix={arXiv},
      primaryClass={cs.CV},
      url={https://arxiv.org/abs/2604.23718}, 
}
\bibliographystyle{ieee_fullname}

\end{document}